\documentclass[sigconf]{acmart}
\usepackage{microtype}
\usepackage{amsfonts} 
\usepackage{pifont}
\usepackage{cancel}
\usepackage[most]{tcolorbox}
\usepackage{soul}
\usepackage{algorithm}
\usepackage{algorithmic}
\usepackage{booktabs}
\usepackage{xspace}
\usepackage{fontawesome}
\usepackage{mathrsfs}
\usepackage{multirow}
\usepackage{amsmath}
\usepackage{colortbl}
\usepackage{subfig}
\usepackage{graphicx}
\usepackage{tabularx}
\usepackage{arydshln}
\usepackage{makecell}
\usepackage{CJKutf8}
\AtBeginDocument{%
  }
\setcopyright{none}

\copyrightyear{2026}
\acmYear{2026}
\setcopyright{cc}
\setcctype{by}
\acmConference[KDD '26]{Proceedings of the 32nd ACM SIGKDD Conference on Knowledge Discovery and Data Mining V.2}{August 09--13, 2026}{Jeju Island, Republic of Korea}
\acmBooktitle{Proceedings of the 32nd ACM SIGKDD Conference on Knowledge Discovery and Data Mining V.2 (KDD '26), August 09--13, 2026, Jeju Island, Republic of Korea}
\acmDOI{10.1145/3770855.3818454}
\acmISBN{979-8-4007-2259-2/2026/08}

\begin{document}

\title{One Model, Multiple Goals: \\Adaptive Multi-Objective Learning for E-commerce Dialogue Systems}





\author{Mingzhe Li}
\affiliation{
  \institution{ByteDance}
  \city{Beijing}
  \country{China}
}
\email{limingzhe.lmz@bytedance.com}

\author{Jing Xiang}
\affiliation{
  \institution{ByteDance}
  \city{Beijing}
  \country{China}
}

\author{Enguo Zhou}
\affiliation{
  \institution{ByteDance}
  \city{Beijing}
  \country{China}
}

\author{Lang Gao}
\affiliation{
  \institution{MBZUAI, ByteDance}
  \city{Abu Dhabi}
  \country{United Arab Emirates}
}

\author{Tai Li}
\affiliation{
  \institution{ByteDance}
  \city{Beijing}
  \country{China}
}

\author{Qishen Zhang}
\affiliation{
  \institution{ByteDance}
  \city{Beijing}
  \country{China}
}

\author{Xiangliang Zhang}
\affiliation{
  \institution{University of Notre Dame}
  \city{Notre Dame}
  \country{USA}
}
\author{Xiuying Chen$^{\dagger}$}
\affiliation{
  \institution{MBZUAI}
  \city{Abu Dhabi}
  \country{United Arab Emirates}
}
\email{xiuying.chen@mbzuai.ac.ae}

\thanks{$\dagger$ Corresponding author.} 


\renewcommand{\shortauthors}{Mingzhe Li et al.}

\begin{abstract}
Dialogue systems in e-commerce scenarios often need to satisfy multiple objectives: accurately reasoning over user profiles (e.g., eligibility, credit limit) to ensure correct decision-making and user state interpretation, while also generating natural and faithful responses.
These goals are complementary but not identical.
In this work, we propose MORE, an adaptive Multi-Objective REinforcement learning framework that jointly optimizes \textit{reasoning accuracy} and \textit{linguistic naturalness}.
Our preliminary experiments show that directly mixing rewards with diverging optimization dynamics can cause oscillations and unstable learning.
Thus, instead of optimizing a single mixed reward, we treat reasoning functions as constraints that guide policy optimization. 
At inference time, the system directly generates responses without explicit reasoning steps, while still benefiting from reasoning-enhanced scaffold and avoiding additional inference overhead.
To better balance linguistic objectives during response generation, we introduce an adaptive multi-reward mechanism that aggregates signals such as fluency and naturalness and dynamically reweighs them via gradient feedback.
We evaluate MORE on two real-world dialogue systems at ByteDance and the MultiWOZ 2.2 benchmark, where it consistently outperforms strong baselines.
In 14-day online experiments on ByteDance production traffic, MORE improves overall and reached conversion by 16.53\% and 30.09\%, while increasing user satisfaction and reducing handoff rates.
Notably, in a human–machine comparison, MORE recovers about 60\% of the incremental conversion lift achieved by human agents.
\end{abstract}





\maketitle

\section{Introduction}

\begin{figure}[t]
    \centering
    \includegraphics[scale=0.45]{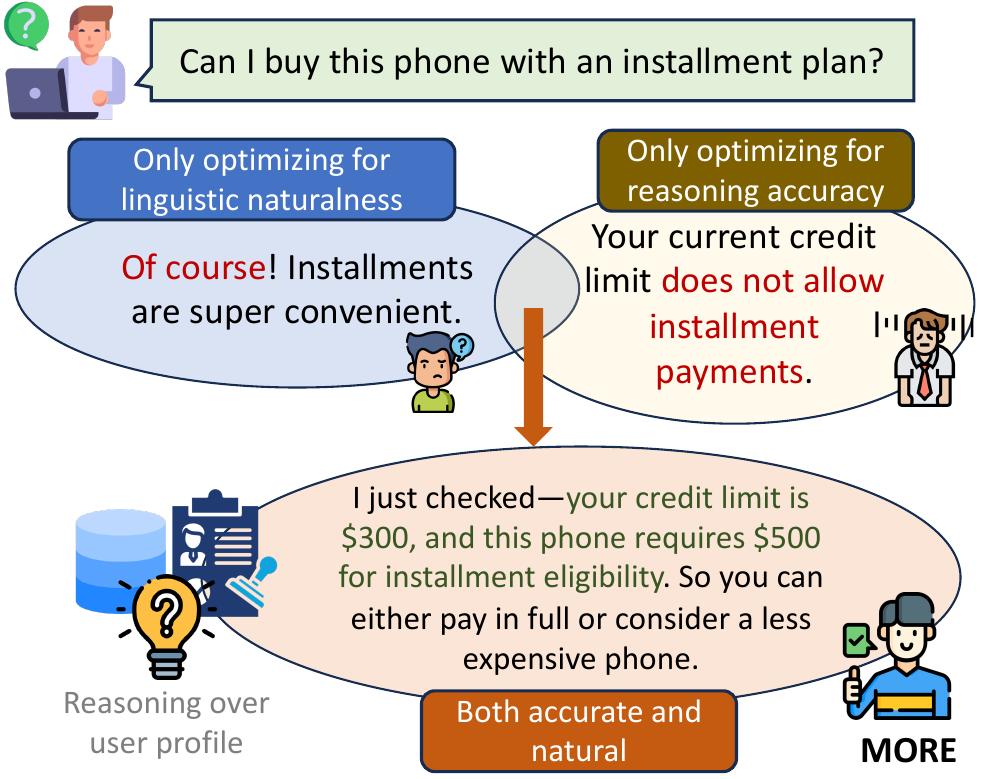}
    \caption{Example of trade-offs in dialogue: fluency-only models sound natural but incorrect; correctness-only models are accurate but rigid. Our model produces responses that are both correct and natural.}
    \label{intro}
\end{figure}

In modern e-commerce ecosystems, dialogue systems are increasingly deployed to handle customer service, product inquiry, and personalized recommendations~\cite{li2025flipping}.
Unlike open-domain chit-chat agents, these systems must satisfy multiple objectives simultaneously.
First, they need to reason over structured user profiles, including eligibility, purchase history, and credit limits, to ensure factual accuracy and trustworthy responses~\cite{song-etal-2025-injecting}.
Second, they must generate responses that are fluent, natural, and semantically faithful, so that users perceive the interaction as both useful and engaging~\cite{see2019makes}. 
These two goals are complementary but not fully aligned.
As shown in Figure~\ref{intro}, a system that excels at reasoning may sound rigid or mechanical, while one that optimizes for conversational quality may risk factual errors or hallucinations.

\begin{figure}[t]
    \centering
    \includegraphics[scale=0.45]{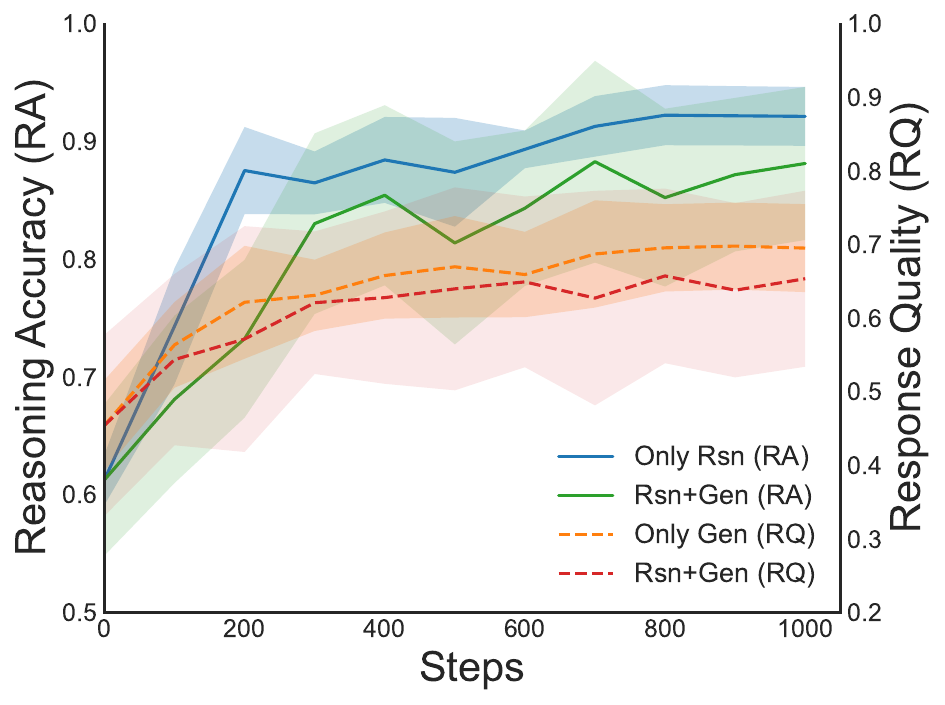}
    \caption{Training dynamics under different reward settings. Shaded regions indicate variance across runs.
Directly combining reasoning accuracy (RA) and response generation (RQ) rewards degrades both metrics and leads to unstable training. }
    \label{dual}
\end{figure}

Existing approaches fall short in different ways. 
Rule-based and symbolic reasoning systems guarantee correctness but often produce stiff, unnatural utterances that hinder user satisfaction~\cite{yang2022interpretable}. 
Another line of research explores personalized dialogue generation, for example, by explicitly modeling user profiles~\cite{song2020generating}.
These approaches primarily aim to improve persona consistency and interpretability, ensuring that the system `sounds like the same person’, but they place little emphasis on factual correctness of user-specific attributes, which is crucial in many real-world scenarios.
There has also been work on reinforcement learning (RL) for task-oriented dialogue, such as hierarchical RL frameworks that separate dialogue policy and natural language generation~\cite{peng2017composite}, and reward function learning for training end-to-end dialogue agents~\cite{feng2022fantastic}.
However, these approaches primarily optimize for task success or slot-filling accuracy, and do not explicitly address the multi-goal nature of dialogue systems, where reasoning correctness and naturalness need to be jointly optimized.
A natural extension is to combine multiple rewards within an RL framework.
Yet, our experiments suggest that directly mixing rewards for reasoning accuracy and linguistic quality often leads to unstable optimization.
As illustrated in Figure~\ref{dual}, naively summing reasoning and generation rewards can degrade both reasoning accuracy and response quality.
In addition to the lower mean performance, the mixed objective exhibits noticeably larger training instability.
Specifically, Rsn+Gen shows substantially higher variance across training steps, with an average standard deviation of approximately 0.07–0.08 for reasoning accuracy and 0.11–0.13 for response quality, compared to 0.02–0.04 and 0.06–0.07, respectively, when optimizing the corresponding objectives alone.
This increased variance indicates stronger oscillations during training and suggests that directly combining heterogeneous rewards leads to unstable optimization dynamics across different cases.
While some instances benefit from joint optimization, indicating that the two objectives are not inherently incompatible, the key challenge lies in how to effectively reconcile and coordinate these competing optimization directions.
We defer a detailed analysis of the underlying mechanisms and extensive empirical results to the Method and Experiment sections.

To address these gaps, we propose MORE, an adaptive Multi-Objective REinforcement learning framework that unifies the optimization of correctness in reasoning over user attributes with response generation quality. 
MORE introduces two core innovations. 
First, we design an auxiliary reasoning refinement to enhance the model’s ability to infer user profiles and attributes. 
Instead of jointly optimizing reasoning and response generation, which may lead to conflicting gradients and oscillation, we use reasoning only as a training scaffold. 
During inference, the model directly generates responses without explicit reasoning, thereby maintaining efficiency while still benefiting from reasoning-aware training.
Second, we introduce an adaptive multi-reward mechanism that aggregates diverse signals, including fluency, naturalness, and semantic fidelity, alongside factual correctness.
Instead of relying on fixed weights, MORE employs a gradient-based dynamic reweighting strategy, enabling the model to flexibly coordinate between competing objectives depending on context.
This allows the system to produce responses that are both trustworthy and natural, a balance that prior single-objective or static multi-reward approaches have struggled to achieve.


To demonstrate the practical significance of the studied problem, we evaluate our approach not only on offline benchmarks but also in real-world production environments. 
Specifically, we deploy our method in large-scale dialogue systems at ByteDance, covering customer service and proactive sales scenarios in the Douyin App. 
The results show that our method improves overall conversion by 16.53\% and reached conversion by 30.09\% in proactive sales, while increasing user satisfaction by 1.12\% and reducing handoff rates to human agents by 1.83\% in customer service. These results demonstrate that effectively balancing multiple dialogue objectives leads to measurable improvements in both dialogue quality and business outcomes in real-world deployments.

Our contributions are threefold. 
Firstly, we identify the fundamental challenge of multi-goal learning in dialogue systems, especially in task-oriented e-commerce scenarios where reasoning correctness and response naturalness must be jointly achieved.
Secondly, we propose MORE, which introduces an adaptive multi-reward mechanism that uses gradient feedback to balance correctness, fluency, and semantic fidelity, while leveraging reasoning as a training  scaffold.
Finally, Experiments in both online and offline settings demonstrate clear gains in response correctness and naturalness over strong baselines.

\section{Related Work}

\textbf{Personalized Dialogue.}
Personalization has long been recognized as a key factor in building effective dialogue systems. 
Early work explored leveraging static user attributes such as age, gender, or preferences to tailor system responses~\cite{li2016persona,zhang2018personalizing}. 
Subsequent studies introduced persona-based datasets and models that explicitly condition generation on user profiles~\cite{see2019makes,chen2022controllable}. 
In e-commerce and customer service scenarios, personalized dialogue is particularly crucial, as users often query about their own accounts, transaction histories, or credit eligibility~\cite{patil2024artificial}. 
However, most existing approaches either focus on shallow personalization (e.g., style or tone) or static attributes, lacking mechanisms for accurate reasoning over numerical or dynamic user information. 
Our work bridges this gap by explicitly enhancing attribute-based reasoning and integrating it into reinforcement learning.

\textbf{Task-Oriented Dialogue.}
Task-oriented dialogue (TOD) systems aim to help users achieve specific goals, such as booking tickets or managing finances, through structured interaction~\cite{xu2024rethinking,li2024multi}. 
Key challenges in TOD include incorporating external knowledge, handling dynamic constraints, and maintaining semantic consistency with user intents~\cite{zhang2020recent,kwan2023survey}. 
While prior TOD research often emphasizes slot filling and database querying~\cite{feng2022fantastic}, little attention has been paid to the unique requirements of e-commerce dialogue~\cite{reddy2024personalization}, which additionally demands reasoning over personalized profiles (e.g., credit limits, interest rates). 
To the best of our knowledge, we are the first to explicitly integrate personalization into reinforcement learning for TOD in e-commerce scenarios.

\textbf{Reinforcement Learning for Dialogue.}
Reinforcement learning has been widely applied to optimize dialogue policies beyond supervised learning~\cite{su2016dialogue, li2016deep}. 
Policy-gradient methods allow models to explore diverse strategies and optimize long-term rewards such as task success, user satisfaction, or diversity~\cite{jaques2019way, zhang2024uoep}. 
More recently, RL from human feedback (RLHF) has proven effective for aligning large language models with human preferences~\cite{yuan2023rrhf,lee2024rlaif}. 
However, standard RLHF typically assumes a single reward signal, which may fail to capture the multiple, and somes conflicting, objectives in dialogue. 
Our MORE framework builds on these advances by introducing adaptive multi-reward optimization with gradient-based reweighting, enabling the model to balance factual correctness and response quality in a unified manner.

\section{MORE}
\begin{figure*}[t]
    \centering
    \includegraphics[scale=0.44]{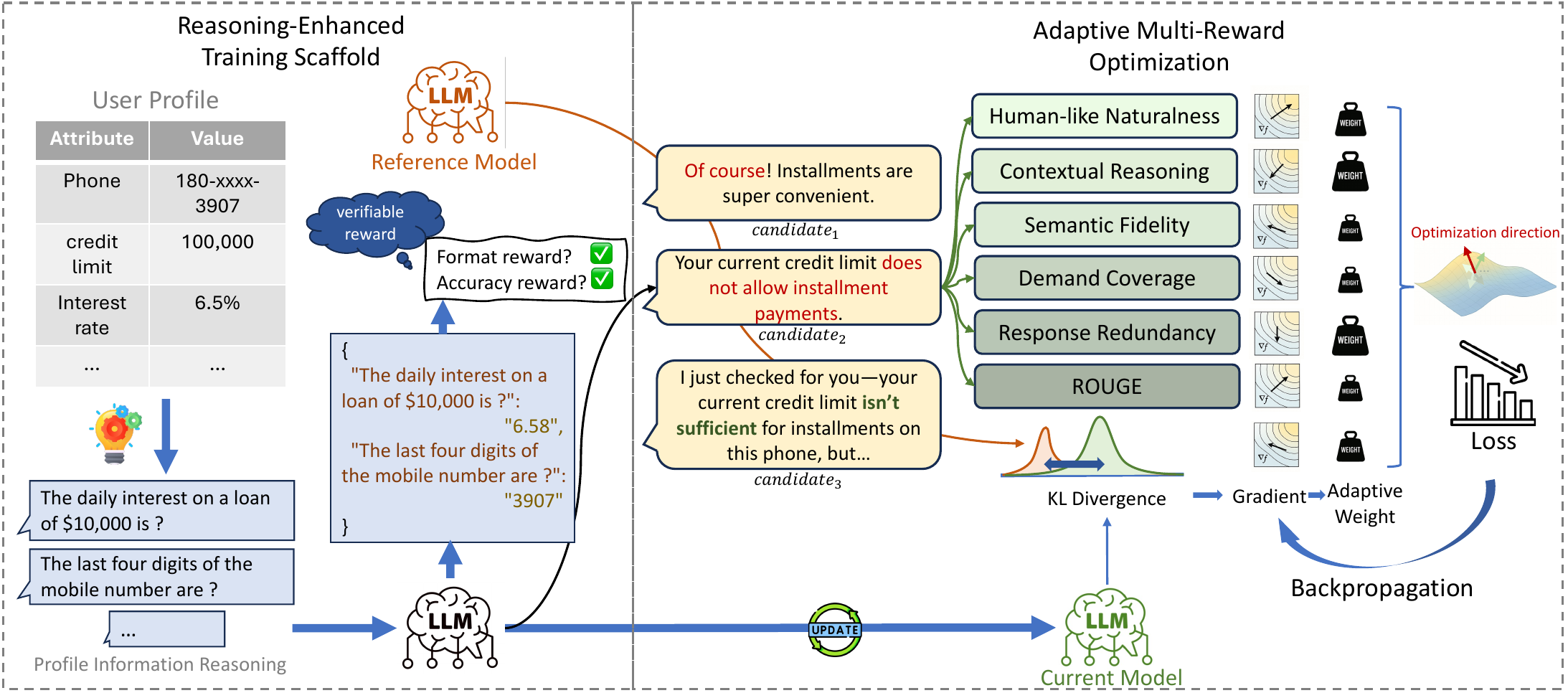}
    \caption{Our MORE framework. 
    A reasoning-enhanced scaffold improves reasoning correctness, while adaptive multi-reward optimization dynamically balances naturalness, reasoning, fidelity, and other objectives to generate responses that are both accurate and natural.}
    \label{model}
\end{figure*}

\subsection{Problem Formulation}
We begin by introducing the notations and key concepts.  
Each user $u$ is associated with a structured attribute database with \(K\) entries $\mathbf{x}_u = \{(a_1, v_1), (a_2, v_2), \ldots, (a_K, v_K)\}$, where the \(k\)-th entry consists of an profile key $a_k$ and its corresponding value $v_k$.
These profiles encode personalized information relevant to the e-commerce scenario, such as credit limit and coupon amount.  
Formally, we model a dialogue as a sequence of $T$ turns $d = \{(x_1, y_1), (x_2, y_2), \ldots, (x_T, y_T)\}$, where $x_t$ denotes the user utterance at turn $t$ and $y_t$ is the corresponding system response.
At each turn, the system generates $y_t$ conditioned on the dialogue history $H_t = \{x_1, y_1, \ldots, x_{t-1}, y_{t-1}\}$, user utterance $x_t$ and the user profile $\mathbf{x}_u$: $y_t = f_\theta(H_t, x_t, \mathbf{x}_u)$.
The quality of a generated response $y_t$ is evaluated against the ground-truth $y_t'$ from two perspectives: (1) \emph{dialogue quality}, assessing fluency and contextual appropriateness over multiple turns; and (2) \emph{reasoning correctness}, assessing whether $y_t$ correctly incorporates or infers information from the user profile $\mathbf{x}_u$.

\subsection{GRPO Preliminary}
\label{subsec:grpo}

Before introducing our adaptive multi-reward optimization, we briefly review the training paradigm of Group-wise Reinforcement Policy Optimization (GRPO) \cite{shao2024deepseekmathpushinglimitsmathematical}, which serves as the backbone of our reinforcement learning stage. 

Concretely, given the dialogue history, user query, and profile, the policy model $f_\theta$ generates a group of $M$ candidate responses $\{y^{(1)}, y^{(2)}, \ldots, y^{(M)}\}$. 
Instead of scoring each candidate independently, GRPO evaluates them jointly within the group. 
This allows the model to learn from relative preferences or normalized rewards, reducing variance compared to traditional per-sample reinforcement learning.

For each candidate $y^{(i)}$, GRPO defines its advantage by normalizing the reward within the group:
\begin{equation}
\label{AA}
A(y^{(i)}) = \textstyle \frac{r_i - \mathrm{mean}(\mathbf{r})}{\mathrm{std}(\mathbf{r})},
\end{equation}
where $r_i = R(y^{(i)})$, $\mathrm{mean}(\mathbf{r})$ is the group mean, and $\mathrm{std}(\mathbf{r})$ is the group standard deviation.

The policy is then optimized using a PPO-style~\cite{schulman2017proximal} objective $\mathcal{L}_{\text{GRPO}}(\theta)$:
\begin{equation}
\label{grpo}
\mathbb{E}\Big[\min\!\big(
r_\theta A,\; g(\epsilon, A)\big)\Big] 
- \beta D_{\mathrm{KL}}(\pi_\theta \,\|\, \pi_{\mathrm{ref}}).
\end{equation}
Here, $r_\theta$ is the importance ratio between the current and old policies, 
$A$ is the group-normalized advantage, 
$g(\epsilon, A)$ denotes the clipped surrogate with threshold $\epsilon$, 
and the last term adds a KL penalty to the reference policy $\pi_{\mathrm{ref}}$ with coefficient $\beta > 0$.


\subsection{Reasoning-Enhanced Training Scaffold}
In our multi-objective reinforcement learning framework, each training step consists of two parts, as shown in Figure~\ref{model}.
The first part is \emph{Profile Information Reasoning}, where we construct a reasoning-enhanced scaffold to explicitly train the model to understand and reason over key user profile attributes, thereby improving both factual recall and numerical inference. 
The second part leverages this reasoning-enhanced signal for \textit{multi-reward optimization}.

\paragraph{\textbf{Profile Database}}
In our e-commerce setting, we construct a profile attribute dataset denoted as $\mathcal{D}_{\text{prof}}$. 
Each entry in $\mathcal{D}_{\text{prof}}$ corresponds to a user $u$ and contains a set of attribute–value pairs 
$\mathbf{x}_u=\{(a_k,v_k)\}_{k=1}^{K}$, 
which encode personalized information such as phone number, ID number, credit limit, and interest rate.

\paragraph{\textbf{From Profile to Training Scaffold}}
Based on $\mathcal{D}_{\text{prof}}$, we design two types of reasoning tasks. 
The first is \emph{Masked Attribute Completion} (MAC), where the user profile is rewritten into natural language contexts and directly retrievable attributes such as the last four digits of a phone number, the credit limit, or the interest rate are masked for the model to recover.
The second is \emph{Programmatic Numerical Reasoning} (PNR), where queries are generated that require numerical inference using profile attributes, such as “What is the daily interest on a 10,000 yuan loan?” or “What is the interest for a $X$ yuan loan over 30 days?”
The former focuses on factual extraction of personalized attributes, while the latter emphasizes reasoning and computation.
We give a full masked example in Appendix~\ref{mask}.

\paragraph{\textbf{Overall Loss Function}}
For MAC tasks, the model predicts the masked attribute value $v_k$ from the user profile. 
For PNR tasks, the model predicts the numerical result derived from the profile through deterministic rules. 
We thus define a reward function $re(y, y^\ast)$ that assigns 1 if the prediction $y$ matches the reference answer $y^\ast$, and 0 otherwise.
The scaffold objective is then optimized via reinforcement learning:
\begin{equation*}
\resizebox{\linewidth}{!}{$
\mathcal{J}_{\text{scaffold}}(\theta) 
= \mathbb{E}_{y \sim \pi_\theta} \left[ 
\lambda_{\text{mac}} \, re_{\text{MAC}}(y, y^\ast) 
+ \lambda_{\text{pnr}} \, re_{\text{PNR}}(y, y^\ast) 
\right]
$},
\end{equation*}
where $\pi_\theta$ is the policy model, and $\lambda_{\text{mac}}, \lambda_{\text{pnr}}$ balance the two reasoning tasks.
This formulation directly optimizes verifiable rewards, encouraging the model to improve factual recall and numerical reasoning ability rather than memorizing answers.
Rather than applying supervised fine-tuning, which often encourages the model to memorize answers, we adopt reinforcement learning to stimulate reasoning ability.

\subsection{Adaptive Multi-Reward Optimization}
\label{multireward}
\paragraph{\textbf{Weakness of direct multi-reward optimization.}}
While the reasoning-enhanced scaffold strengthens the model’s ability to capture user attribute, our ultimate goal also requires fluent and natural responses. 
These two objectives are related but not identical: reasoning ensures correctness, while fluency ensures usability, and directly optimizing both may cause conflicts.
As illustrated in Figure~\ref{dual}, naively summing reasoning and generation rewards degrades both reasoning accuracy and response quality.
Moreover, the mixed objective induces stronger training oscillations and large variance across different cases.
On average, the standard deviation of reasoning performance increases from 0.030 to 0.073 under joint optimization compared to optimizing the reasoning reward alone. 
This substantial increase in variance indicates less stable training dynamics under the mixed objective.
While some instances benefit from joint optimization, indicating that the two objectives are not inherently incompatible, the key challenge lies in how to effectively reconcile and coordinate these competing optimization directions.

This issue is further supported by quantitative results.
As discussed in Section~\ref{ablation} and  shown in Table~\ref{tab:ablation}, directly combining reasoning and generation rewards leads to lower generation quality than training with the generation reward alone.
This suggests that the optimization objective for reasoning can interfere with that of generation, thereby negatively affecting response quality.

Hence, instead of directly optimizing a mixed reward objective, we treat the reasoning-enhanced model as a reference model, denoted as $\pi_{\theta_{\text{ref}}}$ in Eq.~\ref{grpo}.
This reference anchors correctness in personalized reasoning while allowing the policy model to focus on improving response naturalness.

\paragraph{\textbf{Candidate Generation and Reward Computation}}
In each training step, the policy model generates multiple candidate responses $\{y^{(1)}, y^{(2)}, \ldots, y^{(M)}\}$. 
Each candidate is then evaluated along multiple dimensions using a set of reward models or metrics. 
\textit{Human-like naturalness} assesses whether the response is fluent, coherent, and stylistically similar to human dialogue~\cite{liu2021naturalness}. 
\textit{Contextual reasoning consistency} evaluates whether the response is logically aligned with the dialogue history and user profile, while a \textit{repetition penalty} discourages duplication from the previous turn to promote conciseness and diversity~\cite{shi2021refine}. 
\textit{Semantic fidelity} ensures that the response remains faithful to the user utterance and dialogue history, avoiding hallucinations or irrelevant content. 
\textit{Coverage of user demand} measures how well the response addresses explicit or implicit needs, rewarding comprehensive and helpful answers~\cite{deriu2021survey}. 
\textit{Response Redundancy} evaluates the lexical novelty of the response relative to the recent context, specifically overlapping phrasing found within the previous five dialogue turns.
All rewards above are computed using evaluation prompts based on the DeepSeek-R1 framework, ensuring consistent and model-agnostic assessment across dimensions.
The full DeepSeek-R1 evaluation prompts and instructions are provided in Appendix~\ref{prompt}.
Finally, \textit{overlap-based scores} such as ROUGE provide a complementary perspective by comparing generated responses with ground-truth references.

\begin{algorithm}[t]
\caption{MORE-GRPO Training}
\label{alg:more-grpo}
\begin{algorithmic}[1]
\STATE \textbf{Input:} initial policy model $\pi_{\theta_{\text{init}}}$, reward models $r_{\phi}$, clipped surrogate $g$, group size $M$, PPO clip $\epsilon$, KL coefficient $\beta$, number of batches $B$
\STATE \textbf{Output:} updated policy parameters $\theta$

\STATE Initialize policy model: $\pi_\theta \leftarrow \pi_{\theta_{\text{init}}}$

\FOR{$b = 1,\dots,B$}
    \STATE Set $\pi_{\text{old}} \leftarrow \pi_\theta$ \COMMENT{freeze old policy for importance sampling}
    \STATE Reference construction: update $\pi_\theta$ on MAC/PNR tasks; freeze a copy as reference $\pi_{\text{ref}}$.
    \STATE Candidate generation: sample batch $(x^{(b)},\mathbf{x}_u^{(b)})$; generate $M$ responses
           $\{y^{(i)}\}_{i=1}^M \sim \pi_{\text{old}}(\cdot \mid x^{(b)},\mathbf{x}_u^{(b)})$.
    \STATE Multi-reward evaluation: for each $y^{(i)}$, compute reward set
           $(r_1^{(i)},\dots,r_J^{(i)})$ by running $r_{\phi}$.
    \STATE Rank-based aggregation: estimate gradient strength of each reward $r_j^{(i)}$,
           assign rank-based weights $\lambda_j^{(i)}$, and compute aggregated reward
           $R^{(i)}=\sum_j \lambda_j^{(i)} r_j^{(i)}$.
    \STATE Normalize $R^{(i)}$ within group to obtain advantage $A^{(i)}$.
        \STATE Compute ratio $r_\theta^{(i)} =         \pi_\theta(y^{(i)} \mid x^{(b)}) \big/ \pi_{\text{old}}(y^{(i)} \mid x^{(b)})$.
        \STATE Compute surrogate $\mathcal{L}_{\text{clip}}=\min\bigl(r_\theta^{(i)} A^{(i)},\, g(\epsilon,A^{(i)})\bigr)$.
        \STATE Add KL regularization term $\beta\,D_{\mathrm{KL}}(\pi_\theta \,\|\, \pi_{\text{ref}})$.
        \STATE Update $\theta$ by minimizing
               $-\mathbb{E}[\mathcal{L}_{\text{clip}}] + \beta\,D_{\mathrm{KL}}$.
\ENDFOR
\end{algorithmic}
\end{algorithm}


\paragraph{\textbf{Gradient-Based Weight Adaptation}}
In multi-reward settings, a key challenge is that the relative importance of different rewards can vary across input cases and generated samples.
For instance, factual correctness may be crucial when user attributes are involved, while fluency or naturalness may dominate in more open-ended responses. 
This makes static weighting schemes insufficient. 
To address this, we adopt a gradient-based dynamic reweighting strategy that adaptively balances competing objectives.
For each reward $r_j$, where $j=\{1,\ldots,J\}$ indexes the set of reward dimensions, we define a per-reward loss 
\(\mathcal{L}_{r_j}(\theta) = - \mathbb{E}_{y \sim \pi_\theta}[r_j(y)]\), 
and compute its gradient contribution with respect to the model parameters 
$g_j=\nabla_\theta \mathcal{L}_{r_j}(\theta)$. 
The magnitude and direction of this gradient indicate the relative influence of the reward on improving the overall objective. 
Rather than directly normalizing gradient magnitudes, we sort $\{\|g_j\|\}$ in descending order and assign weights according to rank: $\lambda_j = 1 - \textstyle \frac{\text{rank}(j)-1}{n}, \quad j=\{1,\ldots,n\}$,
where $n$ is the number of rewards.  
This scheme focuses on relative importance rather than absolute gradient magnitudes, making it less sensitive to small numerical fluctuations and preventing overly large search ranges.
We adopt rank-based linear decay for two reasons. 
First, rank transformation, widely used in robust statistics and evolutionary optimization, has been shown to stabilize noisy signals by discarding unreliable magnitude information while preserving consistent ordering~\cite{hansen2001completely,hettmansperger2011robust}. 
Second, linear decay provides a simple and assumption-light weighting that avoids overemphasizing the largest gradients or suppressing weaker but still useful reward signals, issues that often arise with softmax or power law scaling~\cite{back1994selective}. 

Finally, the aggregated reward is as: $R(y) =\textstyle \sum_j \lambda_j r_j(y)$,
which is used to compute the group-normalized advantage for the GRPO objective in Eq.~\ref{AA}.
Algorithm~\ref{alg:more-grpo} illustrates the training process.

\section{Experiment}
\label{sec:experiment}

\subsection{Dataset}
To comprehensively evaluate our model, we use two real-world dialogue datasets and one benchmark dataset.

\paragraph{Proactive Sale.}
This dataset comes from a real-world outbound proactive sale scenario at ByteDance, where the objective is to introduce campaigns to users, invite participation, and eventually drive conversions in the Douyin App. The data consists of multi-turn conversations between human agents and users, with sensitive information (e.g., surnames, gender, phone numbers) masked. The raw logs were first cleaned using several LLMs (Qwen, Doubao, DeepSeek), followed by manual annotation. 
The dataset contains 30,000/6,000/6,230 samples for training/validation/testing. The average user query length is 9.31 characters, while the average ground-truth response length is 71.49 characters. 
User input features include masked personal information and campaign-specific attributes (e.g., coupon amount, credit amount, annual interest rate). During the reasoning stage, models must infer information such as phone number suffix, coupon validity, borrowing limit, and interest calculation. In the generation stage, models are required to produce fluent and factually correct responses conditioned on user information and dialogue history.

\paragraph{Customer Service.}
This dataset comes from ByteDance’s real-world customer service scenario, aiming to provide accurate answers and improve user satisfaction. It consists of anonymized multi-turn dialogues with sensitive attributes (e.g., name, ID, phone number) masked. The raw data was preprocessed with Doubao LLM for automatic cleaning and then manually annotated into a structured dataset. 
The dataset is split into 12,000/2,000/2,000 samples for training/validation/testing. The average query length is 9.5 characters, and the average ground-truth response length is 120.4 characters. 
Input features include the user’s current question and financial status (e.g., loan service activation, outstanding payments). The reasoning stage requires models to logically infer service eligibility, coupon qualification, and other conditions. The generation stage requires models to output natural and accurate responses based on user information and dialogue context.

\paragraph{MultiWOZ2.2.} We also adapt our method to the standard task-oriented dialogue task, namely the Multi-Domain Wizard-of-Oz dataset~\cite{budzianowski2018multiwoz}, which is one of the most widely used benchmarks.
MultiWOZ contains over 10k fully labeled human-human conversations spanning multiple domains and topics, with annotations for dialogue belief states and dialogue acts. 


\begin{table*}[t]
    \centering
    \renewcommand{\arraystretch}{1.2}   
    \begin{tabular}{@{}l|ccccccc|ccccccc@{}}
      \Xhline{1pt}
      & \multicolumn{7}{c|}{Proactive Sale} &  \multicolumn{7}{c}{Customer Service} \\
Model & BLEU & RG & NAT & RSN & FID & COV & RUD & BLEU & RG & NAT & RSN & FID & COV & RUD\\
    \hline 
    SFT & 46.55 & 61.42 & 0.77 & 1.20 & 1.13 & 1.43 & 0.46 & 36.97 & 54.55 & 0.78 & 1.18 & 1.21 & 1.36 & 0.52\\
    DPO & 43.35 & 58.34 & 1.16 & 1.37 & 1.15 & 1.49 & 0.75 & 35.31 & 53.19 & 0.97 & 1.29 & 1.17 & 1.42 & 0.77\\
    KTO & 43.99 & 60.17 & 1.04 & 1.38 & 1.12 & 1.47 & 0.71 & 36.64 & 55.24 & 0.94 & 1.25 & 1.22 & 1.41 & 0.73 \\
    RLHF & 44.14 & 61.77 & 0.96 & 1.37 & 1.14 & 1.48 & 0.80 & 37.71 & 56.70 & 1.03 & 1.28 & 1.30 & 1.35 & 0.84 \\
    CoT & 46.24 & 62.70 & 0.79 & 1.41 & 1.10 & 1.45 & 0.77 & 37.58 & 56.61 & 0.89 & 1.31 & 1.28 & 1.38 & 0.81\\
    C-DynaOpt & 46.77 & 62.83 & 1.22 & 1.44 & 1.21 & 1.52 & 0.84 & 37.62 & 56.59 & 1.14 & 1.27 & 1.32 & 1.36 & 0.85\\
    EMORL & 46.51 & 62.14 & 1.29 & 1.42 & 1.23 & 1.51 & 0.88 & 37.85 & 56.80 & 1.20 & 1.28 & 1.25 & 1.40 & 0.90\\
    \hline
    MORE & \textbf{47.49} & \textbf{64.13} & \textbf{1.65} & \textbf{1.55} & \textbf{1.28} & \textbf{1.57} &  \textbf{0.95} & \textbf{38.45} & \textbf{57.22} & \textbf{1.35} & \textbf{1.40} & \textbf{1.39} & \textbf{1.46} & \textbf{0.92}\\
    \Xhline{1pt}
    \end{tabular}
    \caption{Offline evaluation results on ByteDance Proactive Sale and Customer Service datasets. 
    Bold text indicates that our model performs significantly better than the second-best baseline according to a t-test (p < 0.05).
    Metrics include BLEU, ROUGE (RG), Naturalness (NAT), Reasoning (RSN), Fidelity (FID), Coverage (COV), and Redundancy (RUD). MORE consistently outperforms supervised, preference-based, and multi-reward baselines across both scenarios.
    }
    \vspace{-5mm}
    \label{tab:exp_1}
  \end{table*}




\subsection{Comparison Methods}
To comprehensively evaluate our approach, on Bytedance datasets, we compare against the following representative baselines: 
\textit{SFT}: Standard supervised fine-tuning with cross-entropy loss, serving as a strong reference but without explicit alignment.  
\textit{DPO}~\cite{rafailov2023direct}: Preference-based optimization using pairwise preference data for more efficient alignment than RLHF.  
\textit{KTO}~\cite{ethayarajh2024kto}: A preference optimization variant designed for improved sample efficiency and stability.  
Negative samples are derived from \textit{GPT-4o} for contrastive preference pairs.
\textit{CoT}~\cite{wei2022chain}: A finetuned baseline encouraging step-by-step reasoning. 
\textit{RLHF}~\cite{stiennon2020learning}: A three-stage pipeline that first fine-tunes a policy with supervised learning, then trains a reward model on human preferences, and finally optimizes the policy using PPO against the learned reward.
\textit{C-DynaOpt}~\cite{min2024dynamic}: Reinforcement learning with multiple fixed rewards, often struggling to balance objectives.  
\textit{EMORL}~\cite{kong2025emorl}: An ensemble multi-objective RL framework that aggregates specialized models via hidden-state combination.

On the MultiWOZ dataset, we incorporate the performance of both SOTA full-shot training models and zero-shot prompting methods.
The full-shot training models include \textit{SimpleTOD}~\cite{hosseini2020simple}, \textit{UBAR}~\cite{yang2021ubar}, \textit{GALAXY}~\cite{he2022galaxy}, \textit{Mars}~\cite{sun2023mars}, \textit{TOA-TOD}~\cite{bang2023task}, and \textit{RewardNet}~\cite{feng2022fantastic}. 
For prompting-based approaches, we report results from three representative methods. 
\textit{SGP-TOD}~\cite{zhang2023sgp} is a schema-guided prompting method that builds TOD systems with LLMs in a lightweight way. \textit{AutoTOD}~\cite{xu2024rethinking} abandons the traditional pipeline and instead relies on an instruction-following LLM with external APIs. \textit{ProTOD}~\cite{dong2025protod} adopts a progressive prompting framework that refines LLM responses with structured task guidance.
The reported results are taken from ~\cite{dong2025protod}.

\subsection{Implementation Details}
We implemented all experiments in PyTorch, and trained both our model and all baselines under identical settings on eight GPUs. 
Input truncation lengths were set to 1,024, 2,048, and 512 tokens for the Proactive Sale, Customer Service, and MultiWOZ datasets, respectively. 
In MORE, we first rescale each reward signal to the $[0,1]$ interval, and then perform adaptive-weight search and training. 
$\lambda_{\text{mac}}$ and $\lambda_{\text{pnr}}$ are both set to 0.5.
All baselines and our model use Qwen3-8B as the backbone for the Proactive Sale task, Qwen2.5-32B for the Customer Service task, and LLaMA-8B for the MultiWOZ benchmark, all with the same prompt template in Appendix~\ref{prompt}.
MORE is trained for 5 epochs, with a total training time of approximately 47 hours.
Finally, we retain the five best checkpoints based on validation performance and report the average results on the test set.

\subsection{Evaluation Metrics}
For offline evaluation on ByteDance datasets, as introduced in \S~\ref{multireward}, we include two categories of metrics: 
\textit{LLM-based metrics} (Human-like Naturalness, Contextual Reasoning Consistency, Repetition Penalty, Semantic Fidelity, Coverage of User Demand, Response Redundancy,
and \textit{word-overlap metrics} (BLEU, ROUGE).
On the MultiWOZ benchmark, we follow the standard evaluation metrics.
\textit{BLEU}~\cite{papineni2002bleu} measures n-gram overlap with references. 
\textit{Inform} checks if the correct entity is provided, and \textit{Success} verifies if all requested attributes are met.  
\textit{Combined} is computed as BLEU + 0.5 × (Inform + Success).  
We further report language diversity metrics, including conditional bigram entropy (CBE), the number of unique unigrams (\#uniq), and the number of unique trigrams (\#uniq.Tri).

For online evaluation, we adopt task-specific business metrics.  
\textbf{Proactive Sale:} we evaluate \textit{Overall conv.}, which measures whether a user completes a transaction or action in the app on the same day of the interaction (T0), reported as the relative improvement over the baseline. 
We also report \textit{Reached conv.}, defined as the actual T0 conversion rate within the subset of users who were successfully contacted.
The GSB score is computed as: $\text{GSB} =\textstyle  \frac{g + s/2}{g + s + b}$, where $g$, $s$, and $b$ denote the counts of Good, Same, and Bad cases, respectively.  
\textbf{Customer Service:} We evaluate \textit{User Satisfaction}, which reflects users’ subjective satisfaction and is measured via a pop-up rating prompt shown to users after interactions; and \textit{Handoff rate}, the percentage of conversations escalated to human agents (lower is better), and GSB is defined in the same way as above.

\begin{table*}[t]
\centering
\renewcommand{\arraystretch}{1.2} 
\begin{tabular}{@{}l|ccccccc@{}}
  \Xhline{1pt}
  Model & BLEU & Inform & Success & Combined & CBE & \#uniq. & \#uniq.Tri \\
  \hline
  SimpleTOD & 15.0 & 84.4 & 70.1 & 92.3 & - & - & -\\
  UBAR & 17.6 & 83.4 & 70.3 & 94.5 & 2.10 & 478 & 5238 \\
  GALAXY & 19.6 & 85.4 & 75.7 & 100.2 & 1.75 & 295 & 2275 \\ 
  Mars & 19.9 & 88.9 & 78.0 & 103.4 & 1.65 & 288 & 2264 \\
  TOATOD & 17.0 & 90.0 & 79.8 & 101.9 & - & - & - \\
  RewardNet & 17.6 & 87.6 & 81.5	& 102.2 & 1.99	& 423	& 3942 \\
  \hline
  SGP-TOD & 9.1 & 83.9 & 69.9 & 86.0 & - & - & -\\
  AutoTOD & 9.3 & 87.2 & 82.8 & 94.3 & 2.62 & 1722 & 10188\\
  ProTOD & 8.9 & 91.7 & 83.3 & 96.4 & 3.26 & 1951 & 14345 \\
  \hline
  MORE & 18.7 & \textbf{92.5} (+0.9\%) & \textbf{84.8} (+1.8\%) & \textbf{107.35} (+3.8\%) & 2.96 & \textbf{3226} (+65.4\%) & \textbf{25834} (+80.0\%) \\
  \Xhline{1pt}
\end{tabular}
\caption{End-to-end evaluation results on MultiWOZ. 
Bold text indicates that our model performs significantly better than the second-best baseline according to a t-test (p < 0.05).
In parentheses we report the relative improvement of MORE over the best-performing baseline for each metric.
“–” indicates that the corresponding results are not reported by the baseline.}
\label{tab:exp_woz}
\end{table*}

\subsection{Offline Performance}
\label{offline}
Table~\ref{tab:exp_1} summarizes the offline evaluation results on the ByteDance Proactive Sale and Customer Service datasets. 
Across all metrics, MORE achieves the best overall performance compared with supervised, preference-based, reasoning-based, and multi-reward baselines. 
Specifically, MORE consistently improves both automatic metrics, as well as human-aligned metrics including naturalness, reasoning, fidelity, and coverage.
These results demonstrate the effectiveness of combining reasoning-enhanced training with adaptive multi-reward optimization.

In Table~\ref{tab:exp_woz} we show the performance on the public benchmark MultiWOZ dataset.
Our model still demonstrates significantly higher performance than the baselines in terms of both task success and lexical diversity.
We attribute these improvements to the two objectives introduced in GRPO, namely reasoning and generation. 
The reasoning objective encourages the model to produce coherent intermediate reasoning steps, which improves dialogue planning and leads to higher task completion rates. 
The generation objective guides the model to generate more fluent and diverse surface forms, thereby enriching the lexical variety of responses.
Although BLEU and CBE are slightly lower, this trade-off shows that our model emphasizes semantic consistency and diversity, which are more crucial for practical task-oriented dialogue.

\subsection{Online Experiments}

\label{online}
\begin{table}[t]
\centering
\renewcommand{\arraystretch}{1.2} 
\begin{tabular}{@{}l|cc@{}}
  \Xhline{1pt}
  Dataset & Metric & MORE \\
  \hline
  \multirow{3}{*}{Proactive Sale} 
    & Overall conv. ($\uparrow$) & +16.53\% \\
    & Reached conv. ($\uparrow$) & +30.09\% \\
    & GSB ($\uparrow$) & +61.34\% \\
  \hline
  \multirow{3}{*}{Customer Service} 
    & User sat. ($\uparrow$) & +1.12\% \\
    & Handoff rate ($\downarrow$) & -1.83\% \\
    & GSB ($\uparrow$) & +57.0\% \\
  \Xhline{1pt}
\end{tabular}
\caption{Performance improvements of MORE on ByteDance online A/B testing. 
We further analyze Proactive Sale under two outreach strategies (Cohort A and B), and report detailed T0 and T3 conversion results in Section~\ref{online}.}
\vspace{-3mm}
\label{tab:online_exp}
\end{table}

We conducted a 14-day online A/B test in two scenarios, Proactive Sale and Customer Service (Table~\ref{tab:online_exp}).
For the Proactive Sale scenario, the baseline is a rule-based scripted dialogue system, where conversation flows are predefined offline and dialogue branches are selected based on intent classification.
For the Customer Service scenario, the baseline is a retrieval-based system built on predefined FAQ question–answer pairs, which retrieves the most relevant question according to the user query and returns the corresponding answer.
In Proactive Sale, the campaign reached 3.1M users with a 40\% connection rate. 
We evaluate two user cohorts with different outreach strategies. 
Cohort A consists of users who exited the app midway through an operation and were proactively contacted via phone calls, where the reached T0 conversion rate shows a 21.50\% relative improvement over the baseline. 
Cohort B includes proactively contacted users selected based on user profiling and purchase propensity, achieving a 32.61\% relative improvement in T0 conversion, along with an 8.12\% average daily incremental conversion over a 3-day window (T3).
In Customer Service, we observe consistent improvements in user satisfaction and GSB, with a reduction in handoff rate to human agents, demonstrating the effectiveness of our method in real-world business environments.

We further conduct an additional online experiment in the Proactive Sale scenario to directly compare automated outreach with human agents. 
In this experiment, the same candidate list is randomly split into two equal groups, where one group is contacted by human agents and the other by the automated system.
Compared to a no-contact control, the automated system achieves approximately 60\% of the incremental conversion lift delivered by human agents, while reaching about 89\% of the absolute conversion rate achieved by human outreach.
To evaluate response accuracy and naturalness, we conducted simulated phone-call interactions with human participants. Compared to the baseline, our method achieves a 23.24\% relative improvement in accuracy and a 5.0\% improvement in satisfaction.
These results indicate that the proposed system can recover a substantial portion of human-level effectiveness while enabling scalable deployment.

\begin{table}[t]
    \centering
    \small
    \renewcommand{\arraystretch}{1.2} 
    \resizebox{0.48\textwidth}{!}{
    \begin{tabular}{@{}l|cccccc@{}}
      \Xhline{1pt}
Model & BLEU & RG & NAT & RSN & FID & COV \\
    \hline
    MORE & \textbf{47.49} & \textbf{64.13} & \textbf{1.65} & \textbf{1.55} & \textbf{1.28} & \textbf{1.57} \\
    \hline
    w/o adap. & 46.79 & 63.04 & 0.97 & 1.44 & 1.18 & 1.48 \\ 
    w/o stage & 46.73 & 62.87 & 1.22 & 1.39 & 1.14 & 1.56 \\ 
    w/o rsn. & 47.02 & 63.88 & 1.43 & 1.27 & 1.20 & 1.53\\ 
      \Xhline{1pt}
    \end{tabular}
    }
    \caption{Ablation study on Proactive Sale. 
    Bold text indicates that our model performs significantly better than the second-best baseline according to a t-test (p < 0.05).}
    \vspace{-6mm}
    \label{tab:ablation}
  \end{table}

\section{Analysis and Discussion}

\subsection{Ablation Study}
\label{ablation}
We conduct ablation experiments on the Proactive Sale dataset to assess the contribution of each component in MORE (Table~\ref{tab:ablation}). 
Removing adaptive reward reweighting (w/o adap.) causes the largest drop in naturalness and reasoning, underscoring its importance in balancing fluency with factual correctness. Excluding the two-stage design (w/o stage) reduces both naturalness and fidelity, showing that separating reasoning refinement from generation stabilizes training. 
Excluding the reasoning scaffold (w/o rsn.) weakens reasoning ability and fidelity, confirming the value of explicit attribute reasoning as a training signal. 
Interestingly, although removing reasoning leads to a drop in RSN, the overall generation quality is still better than directly merging the two stages, suggesting that the two objectives exhibit conflicting optimization directions and naively combining them can harm response generation.

\subsection{Human Evaluation}
\label{sec:human}

\begin{figure}[tb]
\centering
\includegraphics[scale=0.43]{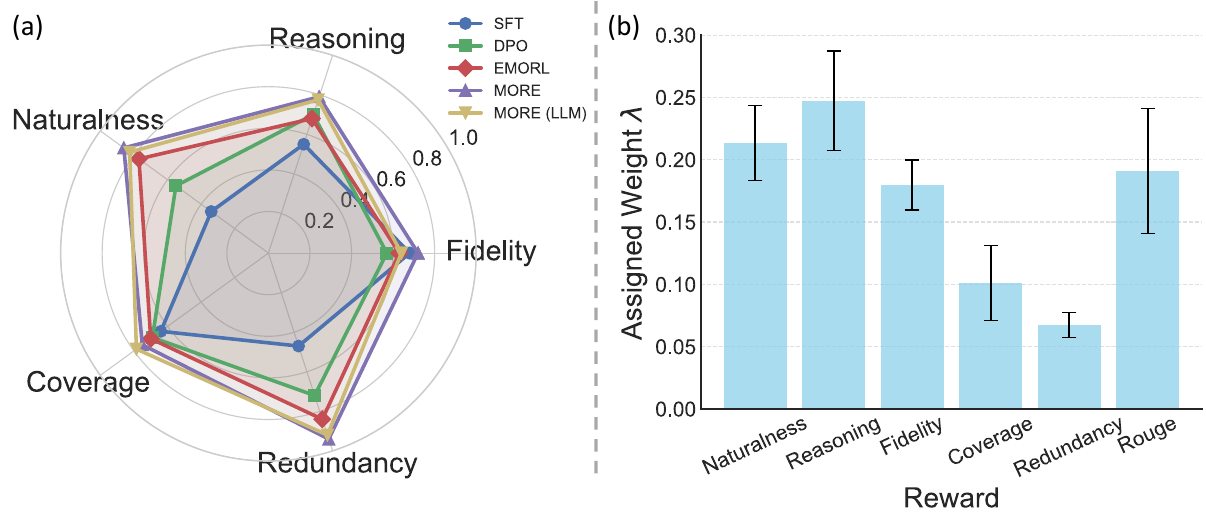}\caption{(a) Human evaluation performance on the Proactive Sale dataset. (b) Distribution of adaptive reward weights, showing context-dependent variation across objectives.}
\label{fig:human}
\end{figure}

To further assess response quality beyond automatic metrics, we conduct a human evaluation on the Proactive Sale dataset.
For each model, 241 samples are randomly selected, and a professional annotation team rates each response as satisfactory (1) or unsatisfactory (0) based on factual correctness, fluency, and helpfulness.
The annotators come from a Chinese content quality and data service platform {\href{https://jobs.bytedance.com/campus/position/7530926574213286151/detail}{\faExternalLink}} responsible for content safety, quality, and user experience across multiple ByteDance products, including Toutiao, Douyin, and Xigua Video, as well as ByteDance commercial content. All annotators are well educated and professionally trained. Inter-annotator agreement, measured by Cohen’s kappa, is 0.66, indicating substantial agreement.

Figure~\ref{fig:human}(a) shows the human evaluation results across Fidelity, Reasoning, Naturalness, Coverage, and Redundancy. 
Overall, MORE achieves the best performance across all metrics, consistently outperforming SFT, DPO, and EMORL, indicating superior response quality from a human perspective.
In addition, the LLM-based evaluation closely matches human ratings, achieving a high Spearman’s rank correlation coefficient ($\rho = 0.77$), which demonstrates strong agreement between automatic and human assessments.


\subsection{Analysis of Adaptive Reward Weighting}
We analyze the distribution of reward weights assigned during training, as shown in Figure~\ref{fig:human}(b).
Overall, the model does not consistently favor a single objective.
Rewards related to linguistic quality, such as naturalness (NAT) and relevance (RSN), receive relatively higher average weights, reflecting their frequent importance in user-facing responses.
Importantly, reward weights exhibit substantial variation across samples, suggesting that the model dynamically adjusts its optimization focus based on input context.
Reasoning-related rewards tend to be prioritized when user attributes or numerical constraints are involved, while naturalness-related rewards dominate in more open-ended interactions.

\subsection{Inference Efficiency and Case Study}
In practical deployment of dialogue systems, inference efficiency is a crucial factor. 
Compared with the COT model that explicitly generates intermediate reasoning steps, our method directly produces the final response during inference. 
As a result, the average latency per sample is reduced from $5.69$ s/item to $1.71$ s/item, achieving a \textit{reduction of 69.9\%}, and the average output length decreases from $252.6$ tokens to $79.8$ tokens, corresponding to a \textit{68.4\% reduction}.
This outcome aligns with our design: reasoning scaffolds are introduced only at the training, while inference omits explicit reasoning, thereby shortening output length, complexity, and time cost.

\begin{figure}[tb]
\centering
\includegraphics[scale=0.46]{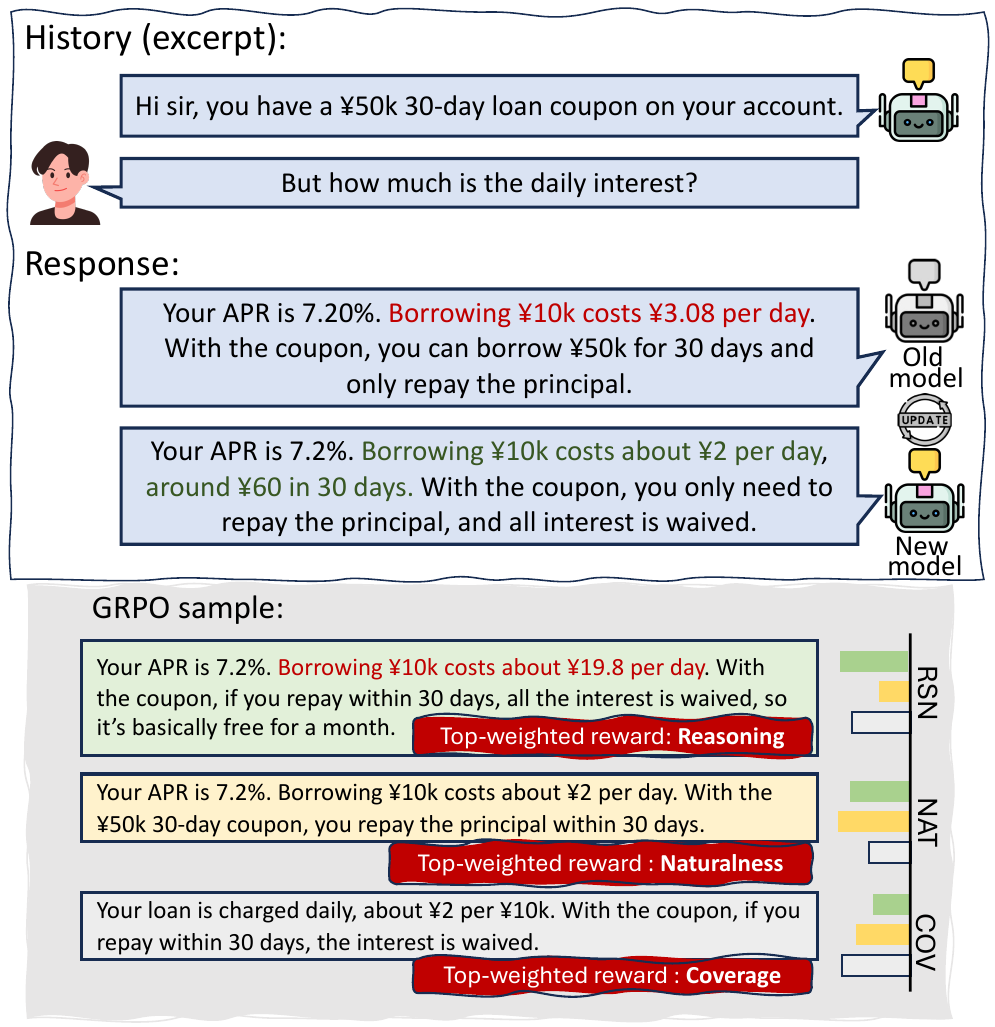}
\caption{Comparison of old vs. new models in a loan interest query, while GRPO samples show how different reward priorities shape responses.
}
\vspace{-3mm}
\label{fig:case}
\end{figure}

We also present a case study from the Proactive Sale dataset to demonstrate our method’s effectiveness in Figure~\ref{fig:case}. 
This case comes from an online loan interest inquiry where the user asked about the cost of a ¥50k 30-day coupon.
The old model showed numerical inconsistencies or unnatural phrasing, whereas the new model combined correct reasoning with fluent expression. 
Then we visualize the adaptive weights assigned to each group sample during GRPO, which highlights how prioritizing different rewards reshapes the response style: Reasoning focus sharpens numerical logic, naturalness focus makes answers more conversational, and coverage focus ensures completeness.
By adaptively reweighting these objectives, MORE flexibly balances accuracy and usability.

\section{Conclusion}
In this paper, we addressed the fundamental challenge of multi-goal optimization in dialogue systems, where factual correctness and conversational naturalness often diverge.
We proposed MORE, an adaptive multi-objective reinforcement learning framework that integrates a reasoning-enhanced training scaffold with gradient-based dynamic reward reweighting. 
Comprehensive experiments across three datasets demonstrate that MORE consistently outperforms diverse baselines, achieving both reasoning accuracy and natural dialogue generation. 
In future work, we plan to extend MORE to broader application domains and investigate its integration with scalable preference learning.

\bibliographystyle{ACM-Reference-Format}
\bibliography{main}

\appendix

\section{Limitations}
While \textsc{MORE} demonstrates strong performance in balancing multiple dialogue objectives, several limitations remain. 
First, our evaluation is limited to Chinese and English task-oriented dialogue scenarios; broader validation on diverse languages and domains is needed to confirm generalizability. 
Second, although dynamic reward reweighting alleviates conflicts between objectives, it relies on predefined reward models whose quality and biases may affect outcomes. 
Third, in the training process, the computational overhead of generating multiple candidates and computing multi-dimensional rewards remains higher than single-objective optimization, which may limit scalability in real-world deployment. 
Finally, our work primarily focuses on offline evaluation; incorporating human-in-the-loop feedback and testing in interactive, real-world settings would provide a more comprehensive assessment.

\section{Ethical Considerations}
Our work focuses on improving task-oriented dialogue systems through adaptive multi-objective reinforcement learning.
While the datasets used in this study are collected from real-world customer service and Proactive Sale scenarios, all sensitive user information (e.g., names, phone numbers, ID numbers) has been strictly anonymized and masked before use.
Data collection and annotation followed internal compliance guidelines to ensure user privacy and security.  

Nevertheless, potential ethical risks remain. 
First, the use of reinforcement learning with multiple reward signals may inadvertently propagate biases from reward models, leading to unequal or stereotypical system behavior. 
Second, optimization for engagement or conversions must be carefully balanced against the obligation to provide transparent, accurate, and user-centered information.  
To mitigate these risks, future deployment of such systems should incorporate fairness-aware reward models, regular auditing, and human oversight. We also emphasize that dialogue agents should be positioned as assistive tools rather than persuasive substitutes, and their usage should remain aligned with ethical standards and regulatory requirements.

\section{Training Sample}
\label{mask}
Here we give a training example. 
The input is a prompt plus masked user information as shown in Figure~\ref{input}.
The output example is in Figure~\ref{output}.

\begin{figure*}[htb]
\begin{tcolorbox}[colback=gray!5!white,colframe=gray!75!black,title={Structured input example}]
Fangxinjie is a loan product provided by Douyin for high-quality customers. Some customers have already been granted credit but have not yet used it. The company is now offering these customers an interest-free coupon that waives loan interest. Your role is a customer service agent of Douyin Wallet Fangxin Loan. Please communicate with the customer based on the reference script and guide them to use the coupon to complete the loan. During this process, if the customer has questions about the product or the company, please answer them based on the corresponding knowledge points. Avoid meaningless pleasantries such as "If you have any questions, feel free to contact me." 

Requirements: (1) Generate the reply based on the reference script, filling in "X" with customer-specific information. The reply must not go beyond the reference script; only minor stylistic adjustments (e.g., interjections, connectors, colloquial expressions) are allowed. (2) Keep the answer concise ($\leq 120$ characters), avoid redundant pleasantries, and do not repeat the conversation summary. (3) Use a colloquial style, addressing the customer with "you," and avoid overly formal written expressions. (4) Adjust wording to avoid repetition with previous responses. (5) Strictly follow the customer information; do not fabricate. The loan amount cannot exceed the credit limit. (6) Convert numbers and symbols into their spoken forms (e.g., 30 days → thirty days, 18.7\% → eighteen point seven percent). 

Below are the telemarketing conversation and the target customer's information. Please generate a reply based on this information, and avoid repeating the agent’s previous turn.

Call information:
Date: 2025-06-03
Customer information: Last name: Guo; Gender: Female; Phone number: 1370887; ID number: 43232*****52X; Credit granted at: 2025-05-08 22:12:17;  Credit limit: 1000.0 CNY; Annual interest rate: 19.8\%Promotion: interest-free coupon for loans within thirty thousand CNY for thirty days, issued on 2025-05-26, valid for seven days, expires on 2025-06-02; History summary of the dialogue; Agent confirmed the customer’s identity and the customer responded to the verification. Historical sequence of dialogue state; Agent confirmed the customer’s identity. Current dialogue turn index 1; Agent’s previous turn; Hello, may I speak with Ms. Guo?; Customer’s current reply; What is this about? Reference script for this turn; We called because we see room to further optimize your rate and limit, so we invite you to participate in our interest-free coupon promotion. You can use X yuan interest-free for X days at no cost. Hello, I am a customer manager for Douyin Wallet’s Fangxinjie, employee ID 95952. I’m calling to inform you about a benefit. We have granted your Douyin Fangxinjie account an interest-free loan coupon that you can use in full; that is, you can withdraw X yuan to your bank card and use it for X days without interest. Whatever you withdraw is what you receive; within X days you only repay principal with no other fees. Are you available now? I’ll take one minute to walk you through it.

Please infer the following user information (phone last digits, ID tail digits, credit limit, Fangxinjie rate, coupon amount, coupon expiry time, interest for borrowing 10,000 CNY for one day, interest for borrowing 10,000 CNY for thirty days), and output it in the JSON format below: \{"phone\_last\_digits": "5266", "id\_tail": "032", "credit\_limit": "130000 CNY", "fangxinjie\_rate": "7.2\%", "coupon\_amount": "50000 CNY", "coupon\_expiry": "May twenty-eighth", "interest\_10k\_1day": "1.97 CNY", "interest\_10k\_30days": "59.18 CNY", "interest\_30k\_1day": "9.86 CNY"\}
\end{tcolorbox}
\caption{An example of the structured input.}
\label{input}
\end{figure*}

\begin{figure*}[t]
\begin{tcolorbox}[colback=gray!5!white,colframe=gray!75!black,title={Structured output example}]
{"phone\_last\_digits": "9876", "id\_tail": "321", "credit\_limit": "1000 CNY", "fangxinjie\_rate": "19.8\%", "coupon\_amount": "30000 CNY", "coupon\_expiry": "June second", "interest\_10k\_1day": "5.42 CNY", "interest\_10k\_30days": "162.74 CNY", "interest\_30k\_1day": "16.27 CNY"}
\end{tcolorbox}
\caption{An example of the generated structured output.}
\label{output}
\end{figure*}

\section{Prompt Template}
\label{prompt}
We provide the detailed prompt templates used for dialogue generation in our experiments in Figure~\ref{prompttt}.

\begin{figure*}[t]
\begin{tcolorbox}[colback=gray!5!white,colframe=gray!75!black,title={Prompt for dialogue generation}]
\{ 'history': [...], \\
\hspace*{0.5cm} `query': 'Fangxinjie is a loan product provided by Douyin for high-quality customers. Some customers have already been granted credit but have not yet used it. The company is now offering these customers an interest-free coupon that waives loan interest. Your role is a customer service agent of Douyin Wallet Fangxin Loan. Please communicate with the customer based on the reference script and guide them to use the coupon to complete the loan. During this process, if the customer has questions about the product or the company, please answer them based on the corresponding knowledge points. Avoid meaningless pleasantries such as "If you have any questions, feel free to contact me." 

Requirements: (1) Generate the reply based on the reference script, filling in "X" with customer-specific information. The reply must not go beyond the reference script; only minor stylistic adjustments (e.g., interjections, connectors, colloquial expressions) are allowed. (2) Keep the answer concise ($\leq 120$ characters), avoid redundant pleasantries, and do not repeat the conversation summary. (3) Use a colloquial style, addressing the customer with "you," and avoid overly formal written expressions. (4) Adjust wording to avoid repetition with previous responses. (5) Strictly follow the customer information; do not fabricate. The loan amount cannot exceed the credit limit. (6) Convert numbers and symbols into their spoken forms (e.g., 30 days → thirty days, 18.7\% → eighteen point seven percent). 

Below is the Proactive Sale scenario with customer profile information. Please generate the dialogue response accordingly.', \\
\hspace*{0.5cm}`response': 'Hello Ms. Guo, this is the customer manager of Douyin Wallet Fangxin Loan, employee ID 95152. I’m calling to share an exclusive benefit: we’ve added an interest-free coupon to your account, allowing loans up to thirty thousand yuan interest-free for thirty days. The full amount will be credited to your bank account, and within thirty days you only need to repay the principal, with no interest or fees. If convenient, I can guide you in the Douyin app to check the coupon—it will only take about a minute.' \}
\end{tcolorbox}
\caption{The detailed prompt template used for dialogue generation.}
\label{prompttt}

\end{figure*}

\end{document}